\documentclass[lettersize,journal]{IEEEtran}
\usepackage{amsmath,amsfonts}
\usepackage{algorithmic}
\usepackage{algorithm}
\usepackage{array}
\usepackage{textcomp}
\usepackage{stfloats}
\usepackage{url}
\usepackage{verbatim}
\usepackage{graphicx}
\usepackage{cite}

\usepackage{diagbox}
\usepackage{multirow}
\usepackage{adjustbox} 

\usepackage{amsmath}
\usepackage{subfigure}
\usepackage{xcolor}
\usepackage[table]{xcolor}

\begin{document}

\title{Targeted Downstream-Agnostic Attack}

\author{Zhuxin Lei, Ziyuan Yang,~\IEEEmembership{Member,~IEEE,}, Liang Liu, Lei Zhang, Jingfeng Lu, Yi Zhang,~\IEEEmembership{Senior Member,~IEEE,}
  \thanks{Zhuxin Lei, Ziyuan Yang, Liang Liu, Lei Zhang, Jingfeng Lu and Yi Zhang are with the College of Computer Science, Sichuan University, Chengdu, China (e-mail: lawliet5697@gmail.com, cziyuanyang@gmail.com, 59154092@qq.com, zhanglei2018@scu.edu.cn, jingfeng.lu@scu.edu.cn, yzhang@scu.edu.cn).}
  }
\markboth{Journal of \LaTeX\ Class Files,~Vol.~14, No.~8, August~2021}%
{Shell \MakeLowercase{\textit{et al.}}: A Sample Article Using IEEEtran.cls for IEEE Journals}


\maketitle

\begin{abstract}
Recently, pre-trained encoders have gained widespread use across various domains due to their strong capability in representation extraction. However, they have been shown to be vulnerable to downstream-agnostic attacks~(DAAs). Existing DAA methods operate under a relatively permissive threat model, where an attack is considered successful if the generated downstream-agnostic adversarial examples~(DAEs) change the original prediction, without requiring a specific target prediction.
In this paper, we propose a novel Targeted DAA~(TDAA) method under a novel and stricter threat model that requires the attack to be both \textit{\textbf{targeted}} and \textit{\textbf{downstream-agnostic}}. To the best of our knowledge, this is the first work to introduce the concept of TDAA.
Since the downstream task is unknown and pre-trained encoders do not directly produce predictions, achieving a targeted attack that forces the model to predict a specific class is particularly challenging. To address this challenge, we rethink the attack pipeline and introduce a novel component termed the “threat image”, which is pre-selected by the attacker as the target. Specifically, a generator is designed to produce example-specific adversarial perturbations that compel the victim encoder to output identical features for both the DAEs and the threat image. Unlike previous DAA methods that generate a single shared adversarial perturbation for all samples, which often fails due to the diversity of images. Our method adopts an example-specific attack paradigm, which generates tailored adversarial perturbations for each image to ensure a high attack success rate and invisibility. By leveraging the threat image as a feature-level anchor, our method builds a task-agnostic bridge to reveal the downstream-agnostic vulnerabilities of the victim encoder. These vulnerabilities are exploited by aligning the latent representations of the DAEs with those of the threat image.
Extensive experiments on 10 self-supervised methods across 3 benchmark datasets demonstrate the effectiveness of our approach and reveal the pronounced vulnerability of pre-trained encoder-based methods, even under our stricter threat model.
\footnote{The code will be made publicly available after the review period.}
\end{abstract}

\begin{IEEEkeywords}
Adversarial robustness, self-supervised learning, pre-trained models, security vulnerability
\end{IEEEkeywords}

\section{Introduction}
\IEEEPARstart{P}{re-trained} encoders, developed using various self-supervised learning (SSL) methods, have become a cornerstone of numerous deep learning~(DL) applications~\cite{clip,sam,GPT}. By leveraging large-scale datasets, these encoders acquire powerful and robust feature representations that are transferable to a wide range of downstream tasks. Hence, using publicly available pre-trained encoders has become a common practice, which allows users to harness knowledge from large-scale datasets to tackle specific tasks more efficiently.

However, similar to other DL methods, pre-trained encoder-based approaches are vulnerable to adversarial attacks~\cite{uap,dae,PAP,ssp}. These methods can be fooled by downstream-agnostic adversarial examples (DAEs), which compromise the performance of downstream tasks without requiring access to either the pre-training or downstream datasets. However, existing downstream-agnostic attack (DAA) methods typically operate under a relatively permissive threat model, which assumes that the primary goal is merely to modify the original predictions without forcing a specific target output. 

Although previous studies~\cite{dae,PAP,uap,uapgd} have demonstrated the vulnerabilities of pre-trained encoder-based models, untargeted attacks often lack practical impact. These attacks lack precision, which limits their effectiveness in exploiting vulnerabilities in a targeted manner. In contrast, targeted attacks can cause greater damage by steering the model towards a specific prediction, posing a more substantial security threat. 


In this paper, we propose a novel and stricter threat model to assess the vulnerabilities of pre-trained encoder-based methods. Specifically, the proposed threat model has two main objectives: 

\textbf{\textit{1) Downstream-Agnostic:}} The attack must remain effective across various downstream tasks, regardless of the specific task or dataset; 

\textbf{\textit{2) Targeted:}} The attack should allow precise manipulation, guiding the model's prediction to a predefined target. 

The first objective has been the exclusive focus of existing DAA methods. In contrast, our threat model retains this basic attack goal while additionally introducing a significantly more challenging attack goal, which extends the baseline threat model adopted in previous works.

Designing an attack method that satisfies both objectives is challenging. Traditional adversarial attack methods typically assume that the downstream task is fixed and known, enabling them to directly leverage the target labels to inject task-specific knowledge into the adversarial perturbation. However, this pipeline is incompatible with our threat model, as we must maintain the downstream-agnostic assumption and ensure the attack’s effectiveness across different downstream tasks, which means we cannot access the label information. This leads to an inherent contradiction between the need for controllability and the requirement of downstream-agnosticism.


To address this challenge, we rethink the attack pipeline and propose the first Targeted downstream-agnostic adversarial attack~(TDAA) method. Specifically, we introduce a threat image in the attack pipeline, which controls the attack by anchoring the generated DAEs to a fixed reference point in the feature space, independent of any downstream task. 

By aligning the features of the generated DAEs with those of the threat image, we can compel pre-trained encoder-based methods to predict the same class as the threat image. This approach eliminates reliance on knowledge of the downstream task, which enables control over the model to produce consistent predictions for both the DAEs and the threat image.

To this end, we design a sample-level DAE generator that produces sample-tailored adversarial perturbations. Unlike previous methods that apply a single perturbation across all samples, our approach generates distinct, example-specific perturbations for each image. Due to the high variability among images, generating a single shared perturbation that achieves both controllability and imperceptibility is challenging. To address this issue, our generator produces customized perturbations for each image. As mentioned earlier, feature alignment between the generated DAEs and the threat image is essential to ensure Targeted predictions. Simultaneously, we require that the degradation in image quality remains within acceptable limits while ensuring control over the model. Therefore, we introduce a quality-preserving loss during training to jointly optimize visual fidelity and attack effectiveness.

Extensive experimental results demonstrate that our TDAA method can successfully attack pre-trained encoder-based methods without access to downstream tasks or datasets, across 10 SSL methods and 3 public datasets. Our method achieves a high targeted fooling rate~(TFR), reaching 100\% in some cases. These findings reveal significant vulnerabilities in existing methods and underscore the necessity of rigorous security evaluation before deployment. The main contributions of this work are summarized as follows:
\begin{itemize}
    \item We propose a novel and stricter threat model for pre-trained encoder-based methods, characterized by two key objectives: downstream-agnosticism and controllability.

    \item Following this strict threat model, we introduce a novel Targeted Downstream-Agnostic Attack~(TDAA) method. To our best knowledge, this is the first Targeted adversarial attack targeting pre-trained encoders.
     
    \item We design a novel sample-level DAE generator that produces tailored adversarial perturbations to individual samples, thereby ensuring attack effectiveness across diverse data. 
\end{itemize}
\section{Related Works}
SSL is capable of leveraging a vast amount of unlabeled data to train a general-purpose encoder, which can be applied to various downstream tasks, such as image classification and object detection. Owing to its ability to overcome the limitations of labeled data, SSL has emerged as a popular paradigm in machine learning. Once pre-trained, the encoder acquires strong feature extraction capabilities. The pre-trained encoder can be directly used for downstream tasks without retraining, which is quite friendly to resource-constrained users. In most cases, only lightweight fine-tuning is required, which incurs minimal computational cost.

Current self-supervised learning techniques are mainly divided into four categories~\cite{ssl-class,ssl-class2}: (1) contrastive learning methods (e.g., MoCo~\cite{Moco2+,Moco3} and SimCLR~\cite{simclr}), the core idea of which is to optimize the model by pulling similar samples closer in the feature space while pushing dissimilar ones farther apart; (2) negative-free methods (e.g., BYOL~\cite{byol} and SimSiam~\cite{simsiam}), which improve representation quality by enforcing consistency between positive samples without relying on negative pairs; (3) clustering-based methods (e.g., SwAV~\cite{swav} and DeepCluster v2~\cite{deepcluster2}), which group similar samples into pseudo-classes using conventional clustering algorithms; (4) redundancy reduction-based methods (e.g., Barlow Twins~\cite{barlow}, W-MSE~\cite{wmse}, and VIbCReg~\cite{vibcreg}), which strengthen intra-dimensional consistency while decorrelating inter-dimensional features to improve representation quality.

\section{Universal Adversarial Examples}
Recent studies~\cite{uap,tfatk2_ttaa,tfatk4_featuremix_cvpr2025,wwwatk1,tdsc_abdukhamidov2023hardening,tdsc_jin2022can} have demonstrated the threat posed by universal adversarial examples to deep neural networks. These examples can greatly interfere with network outputs, and their universal nature allows a single adversarial perturbation to affect multiple samples. Adversarial attacks can be classified into two categories based on their objective:  targeted~\cite{tfatk1_3dtf,tfatk2_ttaa,tfatk3_clip,tfatk4_featuremix_cvpr2025} and untargeted~\cite{dae,uap,uapgd,ssp}. Untargeted attacks only require that the outputs of adversarial examples differ from those of benign examples, while targeted attacks aim to force the outputs of adversarial examples to align with a specific target's outputs. Moreover, with the increasing use of pre-trained encoders, attacks on them have recently emerged~\cite{dae,PAP}. These attacks generate adversarial examples against pre-trained encoders and compromise any downstream models that utilize these encoders. These attacks possess a downstream-agnostic property, and the adversarial examples generated are referred to as DAEs. Current DAEs primarily focus on untargeted attacks, where an attack is considered successful as long as the adversarial example's behavior differs from that of the benign example. To the best of our knowledge, there are currently no DAE-related works on Targeted downstream-agnostic attacks.

\begin{figure}[t]
\centering
\includegraphics[width=\columnwidth]{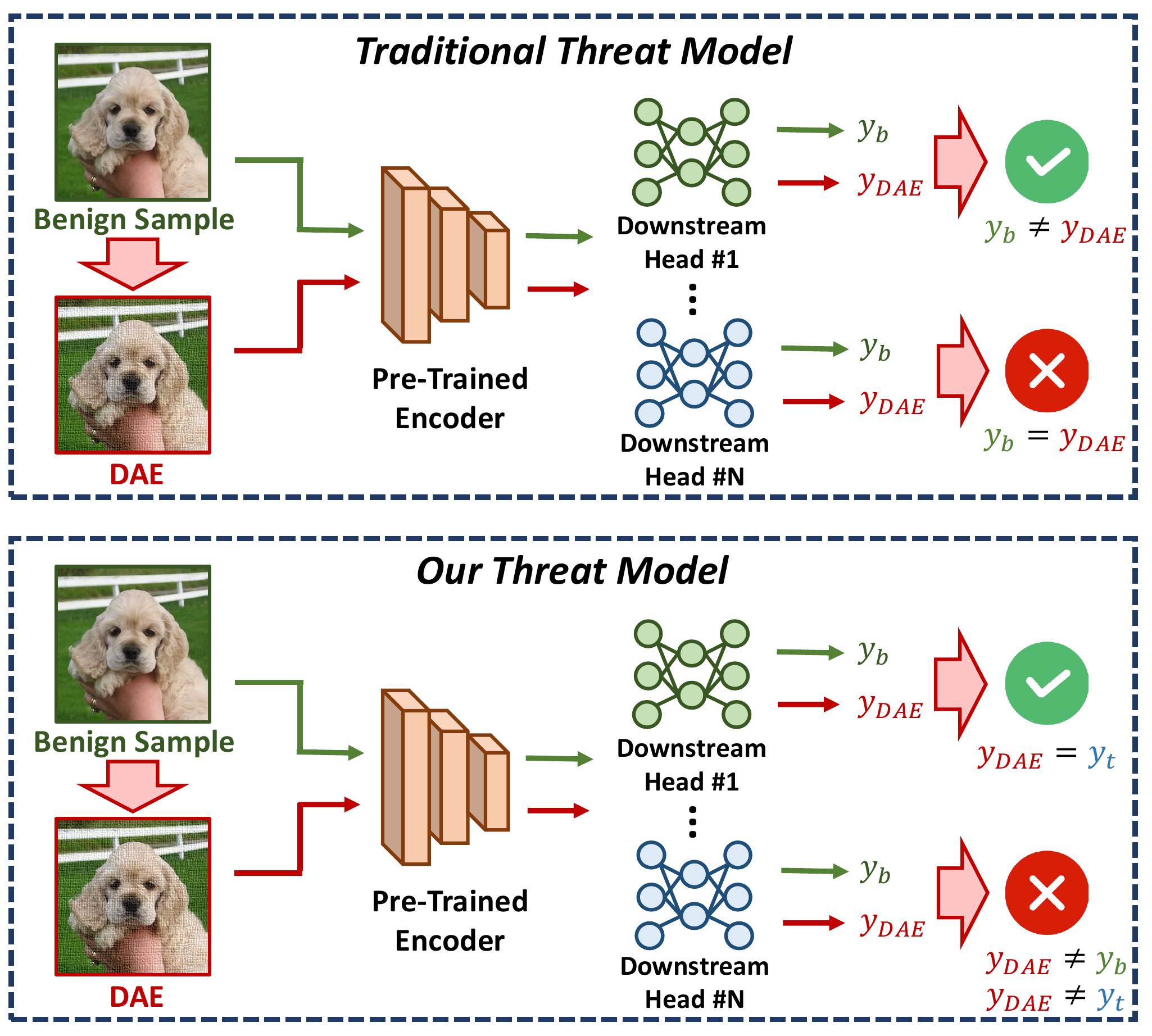} 
\caption{The comparison between the traditional and our proposed threat models.}
\vspace{-10pt}
\label{concept_threat}
\end{figure}

\section{Threat Model}
We assume that attackers can directly download publicly available pre-trained encoders from open platforms, but do not have access to the downstream tasks. The key characteristics can be summarized as follows:

\begin{itemize}
\item \textbf{Universality.} The adversarial examples can attack any downstream model that utilizes the pre-trained encoder, independent of the specific downstream task or dataset.
\item \textbf{Impact.} These adversarial examples effectively degrade the performance of downstream models, with persistence even after fine-tuning or the application of common defense strategies.
\item \textbf{Stealthiness.} The perturbations introduced to the original examples should be imperceptible to humans, ensuring that the attack remains undetected during evaluation or usage.
\item \textbf{Targeted.} The attack must be able to steer the model’s predictions toward a specific target, with predictable and Targeted outcomes that can be forecasted based on the attack design.
\end{itemize}

In the previous works~\cite{dae,uap,PAP,ssp}, only the first three characteristics were considered, which means these works were designed for untargeted attacks. In contrast, in our threat model, even if the predicted result differs from the original one, the attack is still considered a failure unless the predicted label aligns with the expected label. The concepts of these two types of threat models are illustrated in Figure~\ref{concept_threat}.



\begin{figure*}[t]
\centering
\includegraphics[width=\textwidth]{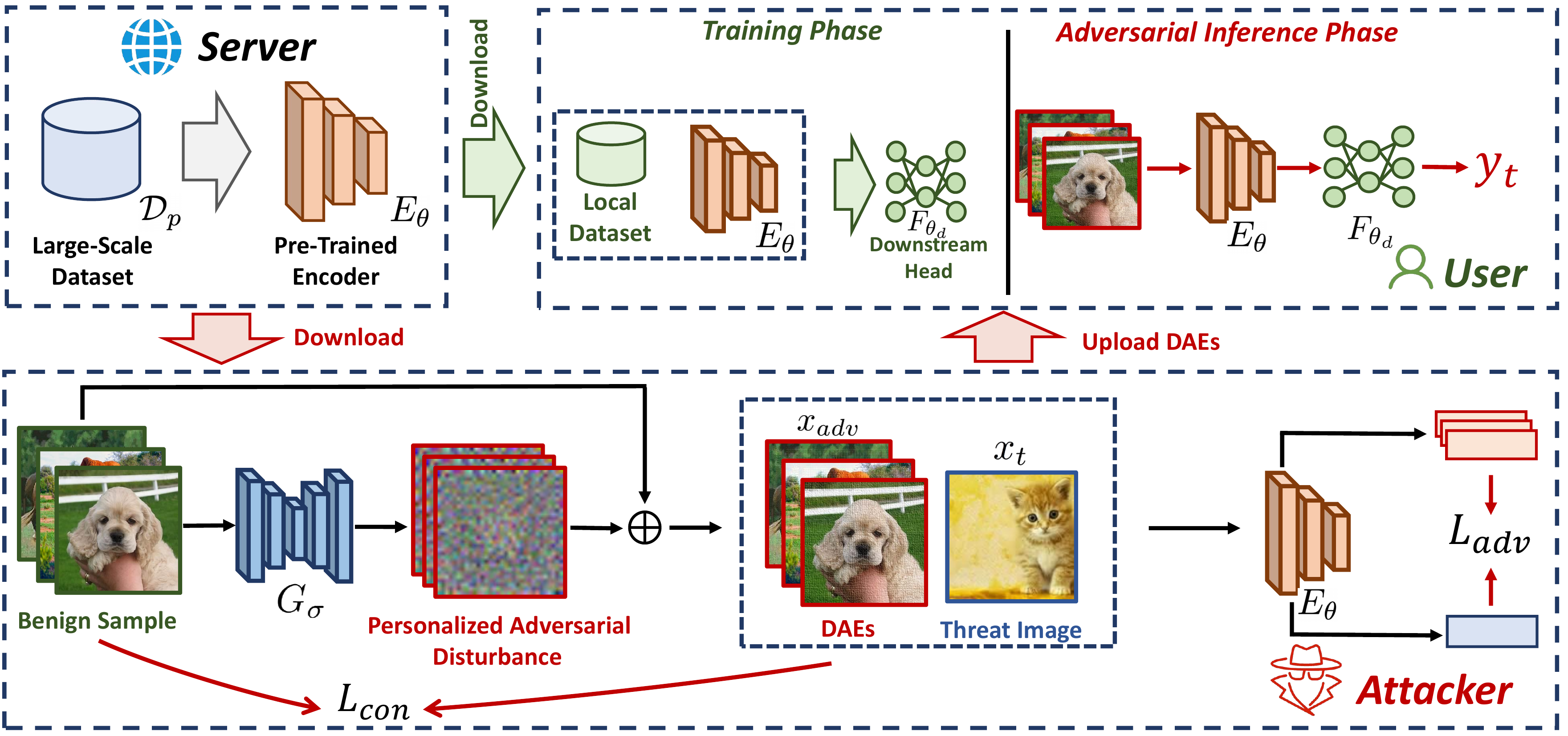} 
\caption{The overview of the proposed TDAA method.}
\label{overview}
\end{figure*}

\section{Methodology}
\subsection{Problem Statement}
Let $E_{\theta}$ denote a powerful pre‑trained encoder parameterized by $\theta$. We define the dataset used to pre‑train $E_{\theta}$ as $\mathcal{D}_{p}$, where each benign example $(x,y)\in\mathcal{D}_{p}$ consists of an input $x$ and its associated label $y$. The feature representation produced by the encoder is given by $v = E_{\theta}(x)$, where $v\in\mathcal{V}$ and $\mathcal{V}$ denotes the feature space. A downstream head is defined as $F_{\theta_{d}}$, parameterized by $\theta_{d}$. Its input is the encoder’s output, so the downstream process can be formulated as $F_{\theta_{d}}(v)$.

In our TDAA, a threat image is selected as the target, represented as $x_{t}$, and we expect to seek a small perturbation $\delta$ to add to the benign input $x$, such that the encoder’s representations of $x_{t}$ and $x+\delta$ become similar. To ensure the perturbation remains imperceptible, $\delta$ is constrained by an upper bound $\epsilon$ based on the $\ell_{p}$-norm. The DAE generation process can be formulated as follows:
\begin{equation}
    \left \|  E_\theta(x_t)-E_\theta(x+\delta ) \right \|_2  \le \eta, \ \text{subject to}\ \| \delta \|_p \leq \epsilon,
\end{equation}
where $\eta>0$ is a small constant that bounds the permitted $\ell_{2}$ distance between the clean and perturbed embeddings.

The attacker’s goal is to fool the downstream classifier $F_{\theta_{d}}$. Specifically, in a classification setting, the attacker aims to induce the same prediction for both the threat image and the generated DAEs. This objective can be formalized as follows:
\begin{equation}
 F_{\theta_{d}}\bigl(E_{\theta}(x_{t})\bigr) \;=\; F_{\theta_{d}}\bigl(E_{\theta}(x + \delta)\bigr), \text{subject to} \| \delta \|_p \leq \epsilon.  
\end{equation}

\subsection{Adversarial Noise Generator Training}
Unlike previous methods that rely on shared perturbations, TDAA introduces a example-specific adversarial perturbation generator. An overview of this approach is provided in Figure~\ref{overview}. Due to the variability in images, a shared perturbation cannot deceive the system across all samples. As a result, the core of our generator is to generate example-specific perturbations for different samples, which ensures that the generated DAEs are both targeted and imperceptible.



Our framework comprises three components: an adversarial noise generator $G_{\sigma}$, a threat image $x_t$, and a publicly downloaded victim encoder $E_{\theta}$. $G_{\sigma}$ is trainable, while the parameters of the victim encoder remain fixed. Specifically, both the user and the attacker share an open-source pre-trained encoder, which is publicly available after being trained on a large-scale dataset. To generate adversarial examples, we feed benign samples from the attacker's dataset \(D_a\) into the generator to produce personalized adversarial noise. Then, the generated noise perturbation is added to the benign samples to generate the DAE \(x_{\text{adv}}\).

As mentioned earlier, the generated DAEs must be both targeted and imperceptible. Therefore, our optimization objectives consist of two main components, and the overall loss function is formulated as follows:
\begin{equation}
    L_{G_{\sigma}}= \alpha L_{adv}+L_{con},
    \label{eq:all_loss}
\end{equation}
where $L_{adv}$ denotes the feature-level adversarial loss function used to control downstream predictions, and $L_{con}$ represents the visual-level consistency loss function designed to preserve image quality. $\alpha$ denotes the tradeoff hyperparameter, which is empirically set to 2 in this paper.

\noindent \textit{\textbf{Feature-Level Adversarial Loss:}} Although the downstream task is unknown, if we can ensure that the features of the generated DAEs align with those of the threat image, we can control the downstream prediction. To achieve this, we guide our generator to produce adversarial noise that transforms benign samples into adversarial examples with feature representations similar to the anchor feature through a loss function. Specifically, we quantify the similarity between the features using the L2 distance. Thus, the feature-level adversarial loss $L_f$ can be defined as:
\begin{equation}
    L_{adv}=\left \| E_{\theta}(x_{adv})-E_{\theta}(x_{t}) \right \|_2,
\end{equation}
where $x_{adv}$ denotes the generated adversarial example, obtained by combining the benign sample $x \in D_{a}$ and the personalized noise generated by $G_{\sigma}$. This process can be formulated as follows:
\begin{equation}
    x_{adv}=x+G_{\sigma}(x).
\end{equation}

\noindent \textit{\textbf{Visual-Level Consistency Loss:}}
To preserve the imperceptibility of the generated DAEs, we introduce the visual-level consistency loss $L_{con}$, which constrains the magnitude of the adversarial noise produced by the generator. Additionally, after each optimization step, we apply a clipping operation to $\delta$ to ensure it satisfies the constraint $\epsilon$. Formally, the process can be formulated as follows:
\begin{equation}
    L_{con}=\left \| x_{adv}-x \right \|_2. 
\end{equation}

\begin{figure}[t]
\centering
\includegraphics[width=\columnwidth]{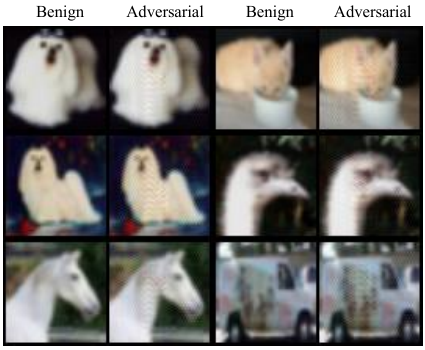} 
\caption{Examples of the benign samples and their corresponding generated DAEs.}
\label{be_vs_adv}
\end{figure}

Finally, the weighted sum of $L_{adv}$ and $L_{con}$ is computed with a tradeoff hyperparameter $\alpha$ to form the final loss function, as shown in Eq.~\eqref{eq:all_loss}. This loss is used to optimize our example-specific adversarial noise generator, $G_\sigma$. In this way, our proposed method generates DAEs with limited disturbance, and some examples are shown in Figure~\ref{be_vs_adv}.


\section{Experiments}
\subsection{Experimental Setting}
Following previous works~\cite{dae,tfatk2_ttaa}, we selected CIFAR10~\cite{cifar10}, STL10~\cite{stl10}, and ANIMALS10~\cite{animals10} as the evaluation datasets, and used ResNet as the backbone for the noise generator. For the pre-trained models, we utilized pre-trained encoders from the publicly available SSL library, solo-learn, as the victim encoders. All encoders were pre-trained on CIFAR10, with ResNet18~\cite{resnet} serving as the backbone. Ten SSL methods were selected to evaluate our performance (BYOL~\cite{byol}, W-MSE~\cite{wmse}, SimCLR~\cite{simclr}, MoCo v2+~\cite{Moco2+},MoCo v3~\cite{Moco3}, NNCLR~\cite{nnclr}, ReSSL~\cite{ressl}, SupCon~\cite{supcon}, SwAV~\cite{swav}, and VibCreg~\cite{vibcreg}). 

\subsection{Performance Metric}
\subsubsection{Targeted Fooling Rate.}
Targeted Fooling Rate (TFR) reflects the proportion of DAEs classified by the model as the target class $y_t$. TFR can be computed as follows:
\begin{equation}TFR=\frac{\sum_{i=1}^{N}g(y_{DAE}^i,\ y_t)}{N},g(a,b)=\left\{
\begin{aligned}
1 ,  a=b \\
0 ,a\ne b 
\end{aligned}
\right.
\end{equation}
where \( N \) represents the number of DAEs, \(y_{DAE}^i\) is the predicted label of the adversarial example, and \(y_t\) denotes the predicted label of the target example chosen by the attacker.

\subsubsection{Adversarial Test Accuracy.}
Adversarial Test Accuracy~(ATA) refers to the accuracy with which downstream models correctly classify the corresponding benign label for the generated DAEs. A lower ATA indicates better attack performance.

\begin{table*}[]
\caption{The TFRs (\%) and ATAs (\%) of TDAA under different settings.\( D_{a} \) represents the dataset used by the attacker, while \( D_{d} \) denotes the dataset used for downstream tasks.}
\label{basic}
\resizebox{\textwidth}{!}{
\begin{tabular}{ccccccccccccc|c}
\hline
$D_a$                      & $D_d$                      & Metric         & BYOL   & W-MSE & SimCLR & MoCo2+ & MoCo3 & NNCLR  & ReSSL  & SupCon & SwAV   & VibCreg & Average    \\ \hline
\multirow{6}{*}{CIFAR10}   & \multirow{2}{*}{CIFAR10}   & ATA$\downarrow$ & 10.21  & 10.21 & 10.21  & 10.21  & 10.21 & 10.21  & 10.21  & 10.21  & 10.21  & 10.21   & 10.21  \\
                           &                            & TFR$\uparrow$  & 100.00 & 99.99 & 100.00 & 100.00 & 99.98 & 100.00 & 100.00 & 100.00 & 100.00 & 100.00  & 100.00 \\ \cline{2-3}
                           & \multirow{2}{*}{STL10}     & ATA$\downarrow$ & 16.70  & 28.59 & 9.99   & 10.42  & 32.28 & 10.07  & 10.56  & 9.99   & 9.99   & 9.99    & 14.86  \\
                           &                            & TFR$\uparrow$  & 88.18  & 73.75 & 100.00 & 99.48  & 73.39 & 99.80  & 98.93  & 100.00 & 100.00 & 100.00  & 93.35  \\ \cline{2-3}
                           & \multirow{2}{*}{ANIMALS10} & ATA$\downarrow$ & 36.74  & 27.05 & 8.04   & 14.42  & 55.39 & 28.49  & 21.92  & 7.58   & 9.40   & 15.48   & 22.45  \\
                           &                            & TFR$\uparrow$  & 33.65  & 54.44 & 97.43  & 26.72  & 21.20 & 39.98  & 67.03  & 98.25  & 95.68  & 73.93   & 60.83  \\ \hline
\multirow{6}{*}{STL10}     & \multirow{2}{*}{CIFAR10}   & ATA$\downarrow$ & 10.21  & 13.36 & 10.21  & 10.21  & 10.25 & 10.21  & 10.21  & 10.21  & 10.21  & 10.21   & 10.53  \\
                           &                            & TFR$\uparrow$  & 100.00 & 93.17 & 100.00 & 99.98  & 99.95 & 100.00 & 100.00 & 100.00 & 100.00 & 100.00  & 99.31  \\ \cline{2-3}
                           & \multirow{2}{*}{STL10}     & ATA$\downarrow$ & 9.99   & 31.95 & 9.99   & 10.11  & 13.79 & 9.99   & 9.99   & 9.99   & 9.99   & 9.99    & 12.58  \\
                           &                            & TFR$\uparrow$  & 100.00 & 55.75 & 100.00 & 96.47  & 95.64 & 100.00 & 100.00 & 100.00 & 100.00 & 100.00  & 94.79  \\ \cline{2-3}
                           & \multirow{2}{*}{ANIMALS10} & ATA$\downarrow$ & 6.98   & 24.93 & 6.53   & 14.22  & 21.05 & 7.35   & 6.87   & 6.61   & 6.62   & 6.56    & 10.77  \\
                           &                            & TFR$\uparrow$  & 99.16  & 74.71 & 99.90  & 51.00  & 65.63 & 99.07  & 98.67  & 98.46  & 99.84  & 99.88   & 88.63  \\ \hline
\multirow{6}{*}{ANIMALS10} & \multirow{2}{*}{CIFAR10}   & ATA$\downarrow$ & 10.21  & 11.51 & 10.21  & 10.21  & 13.96 & 10.21  & 10.21  & 10.21  & 10.21  & 10.21   & 10.72  \\
                           &                            & TFR$\uparrow$  & 100.00 & 98.46 & 99.83  & 100.00 & 96.01 & 100.00 & 100.00 & 100.00 & 100.00 & 100.00  & 99.43  \\ \cline{2-3}
                           & \multirow{2}{*}{STL10}     & ATA$\downarrow$ & 10.36  & 26.45 & 9.99   & 10.39  & 29.64 & 9.99   & 9.99   & 9.99   & 9.99   & 9.99    & 13.68  \\
                           &                            & TFR$\uparrow$  & 99.48  & 77.99 & 99.96  & 99.51  & 77.38 & 100.00 & 100.00 & 100.00 & 100.00 & 100.00  & 95.43  \\ \cline{2-3}
                           & \multirow{2}{*}{ANIMALS10} & ATA$\downarrow$ & 6.61   & 13.61 & 6.48   & 6.50   & 10.29 & 6.48   & 6.48   & 6.51   & 6.48   & 6.48    & 7.59   \\
                           &                            & TFR$\uparrow$  & 99.86  & 86.45 & 100.00 & 99.98  & 91.19 & 100.00 & 100.00 & 99.95  & 100.00 & 100.00  & 97.74  \\ \hline
\end{tabular}
}
\end{table*}


\subsection{Attack Performance}
\textbf{Implementation Details.} To demonstrate the downstream-agnostic characteristic, we designed experiments where the attacker's dataset and the downstream task dataset were inconsistent. During training, we set the hyperparameter \(\alpha\) to 2. The initial learning rate was set to 0.0002, utilizing the Adam optimizer~\cite{kingma2014adam}.

\noindent \textbf{Analysis.} Our experimental results demonstrate the attack effectiveness of TDAA and the vulnerability of pre-trained encoder-based methods. As shown in Table~\ref{basic}, TDAA performs effectively across nearly all downstream tasks, with TFR exceeding 95\% in most cases. Moreover, our method can achieve 100\% TFR in certain settings.

\begin{figure}[!t]
\centering
\subfigure[Benign Samples' features]{\label{fig:subfig:a}
\includegraphics[width=0.48\linewidth]{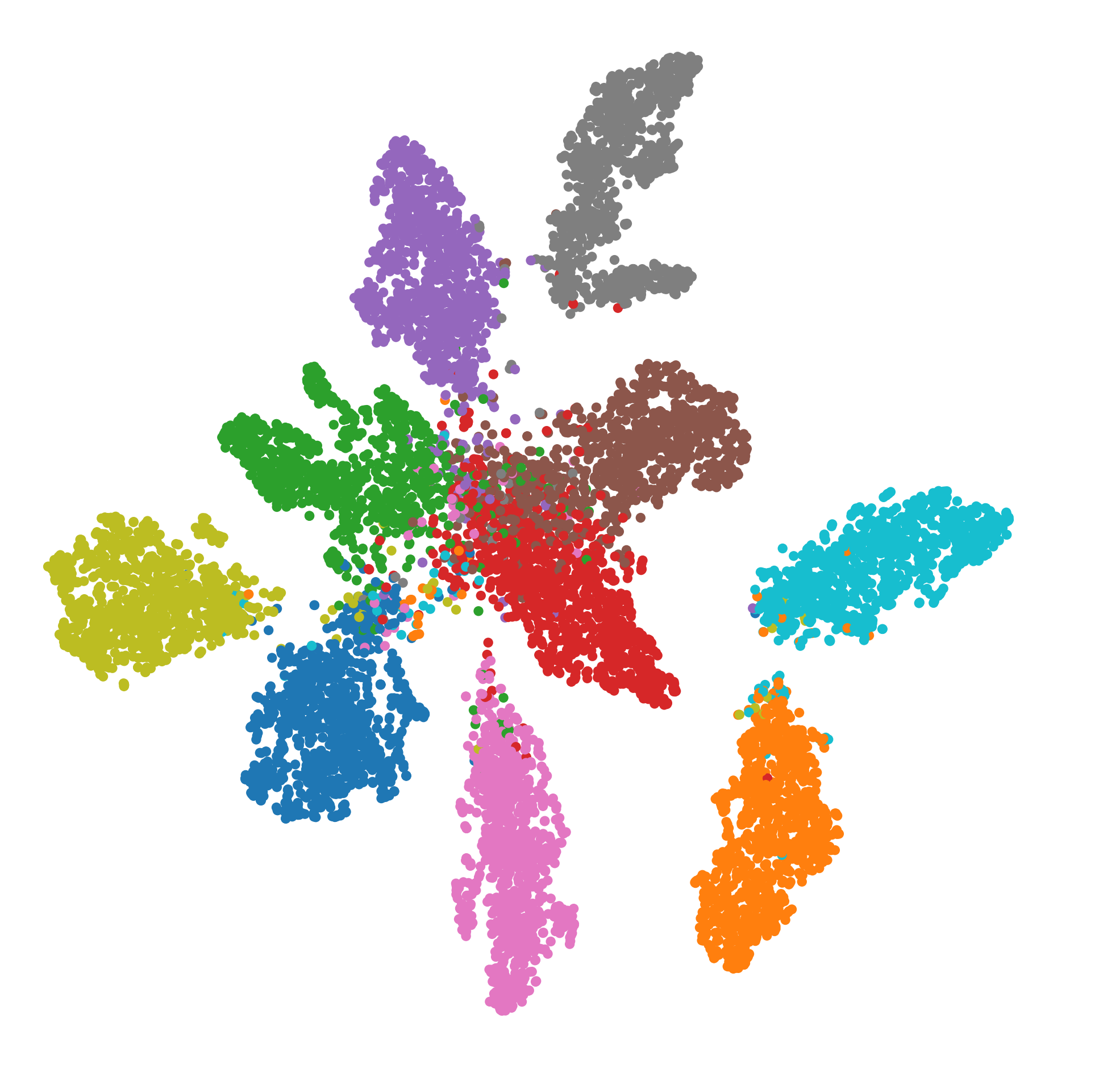}}
\subfigure[DAEs' features]{\label{fig:subfig:b}
\includegraphics[width=0.48\linewidth]{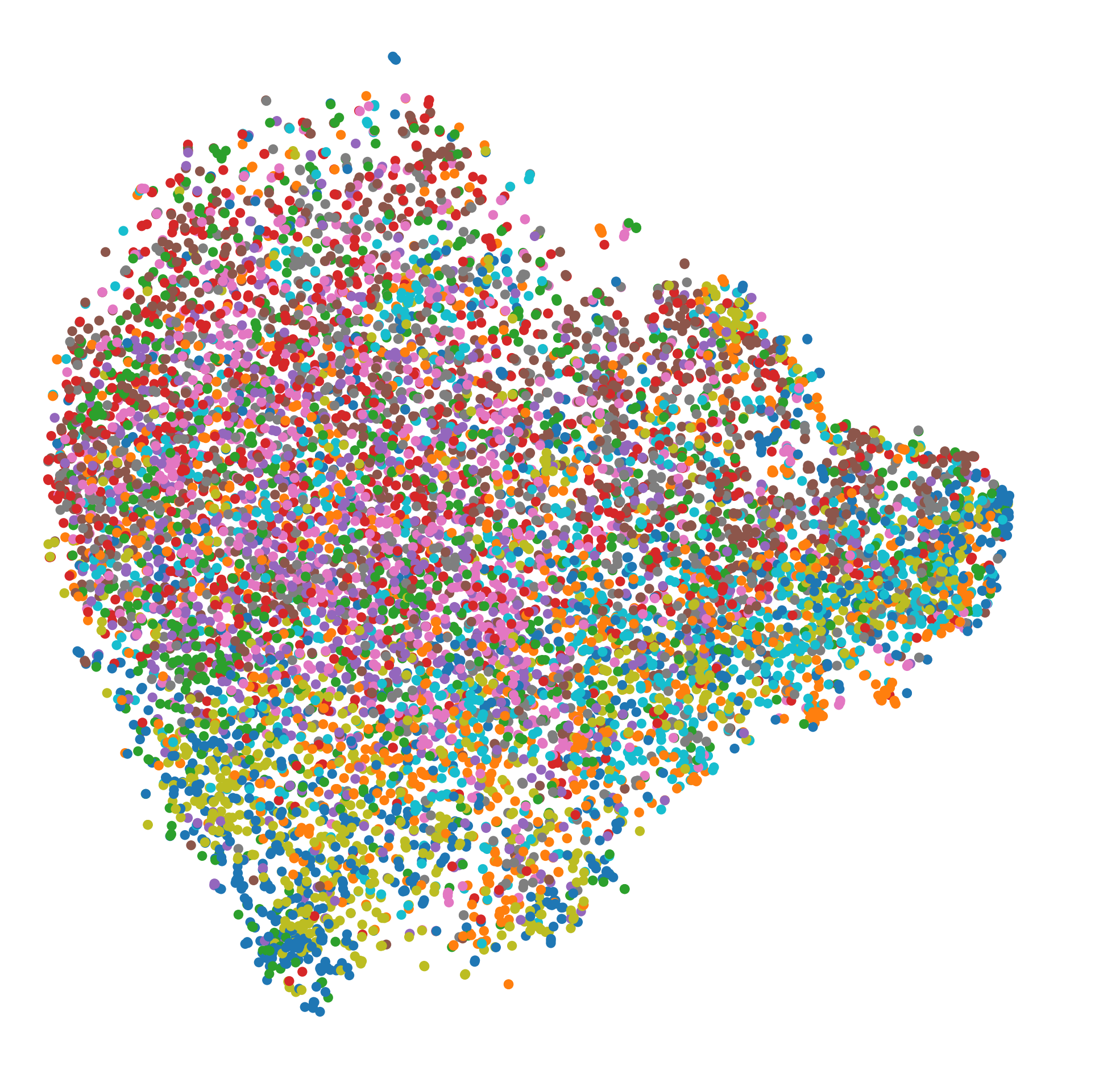}}
\subfigure[Benign Samples' logits]{\label{fig:subfig:a}
\includegraphics[width=0.48\linewidth]{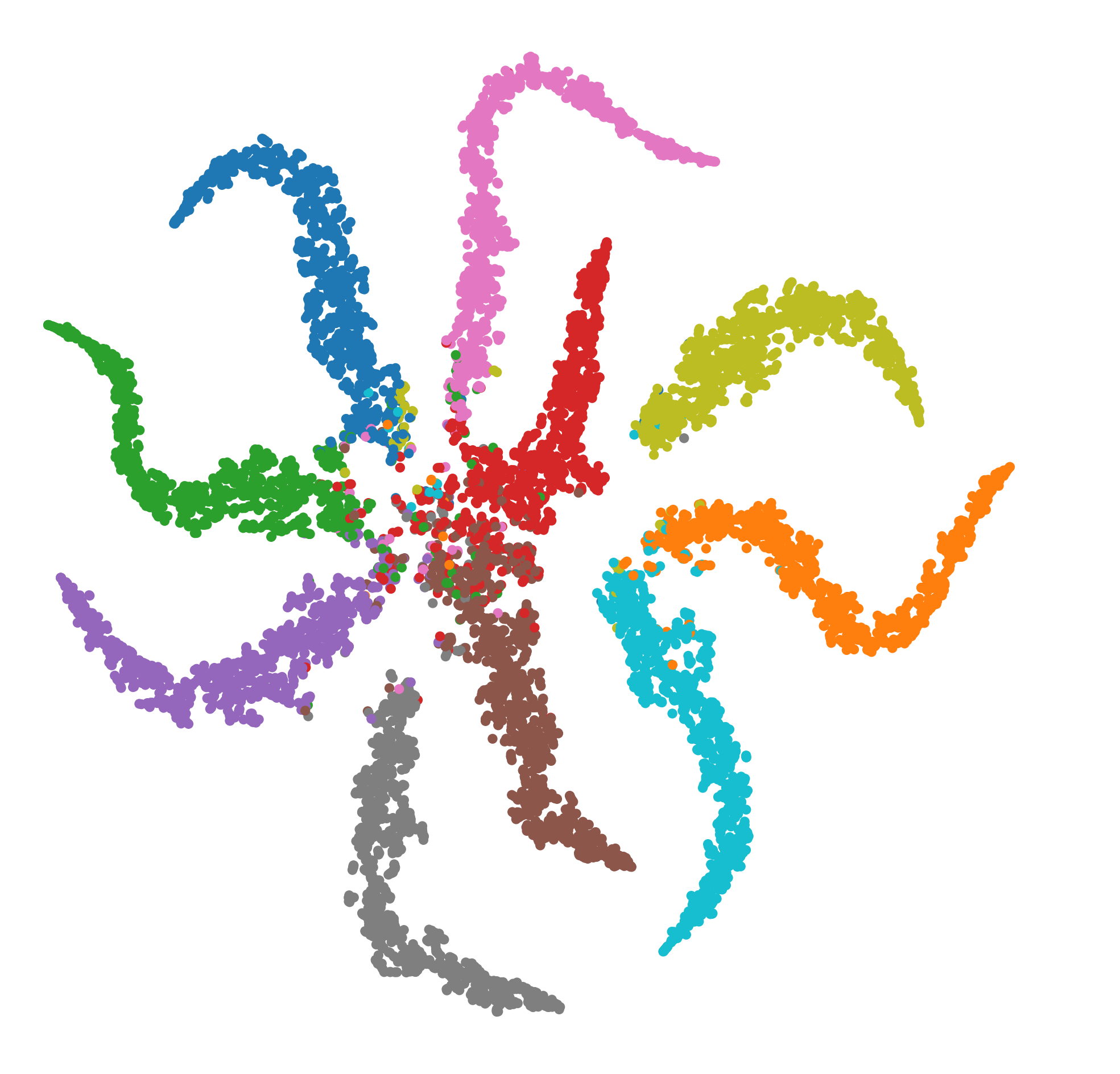}}
\subfigure[DAEs' logits]{\label{fig:subfig:b}
\includegraphics[width=0.48\linewidth]{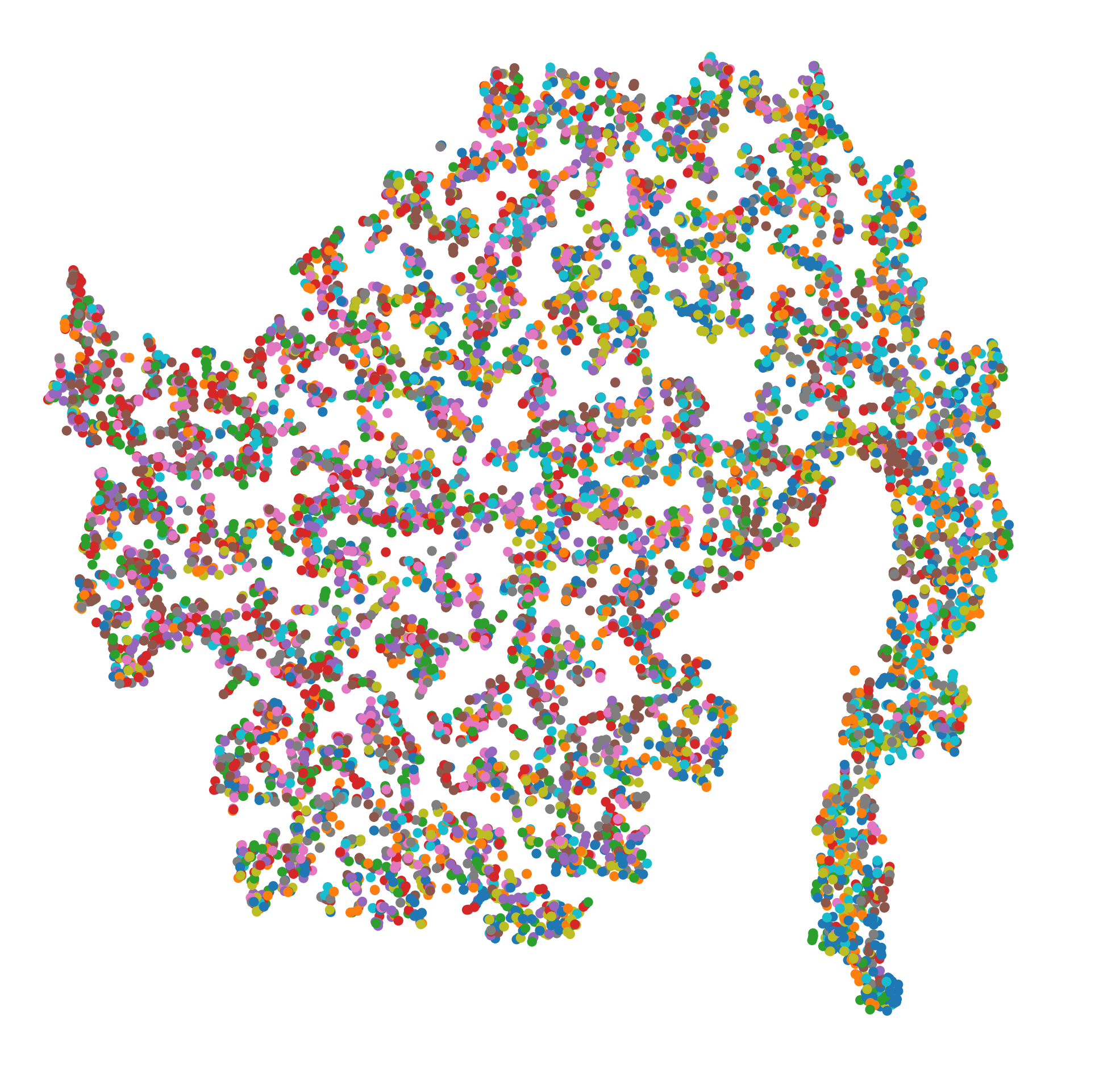}}
\caption{The t-SNE visualization results illustrate the distribution of DAE and benign samples in both the feature space and the logits space.}
\label{figtsne}
\end{figure}

To provide an intuitive understanding of the effectiveness of TDAA, we present t-SNE visualizations in Figure~\ref{figtsne}. Specifically, we visualize the features and logits of benign and adversarial examples using the CIFAR10 downstream dataset. In the absence of our attack, both the features and logits are clearly and correctly separated into different clusters. However, the logits and features of our generated DAEs cannot be distinguished, which demonstrates that our method effectively performs the attack and confuses the model's predictions.

\begin{figure*}[t]
\centering
\includegraphics[width=.95\textwidth]{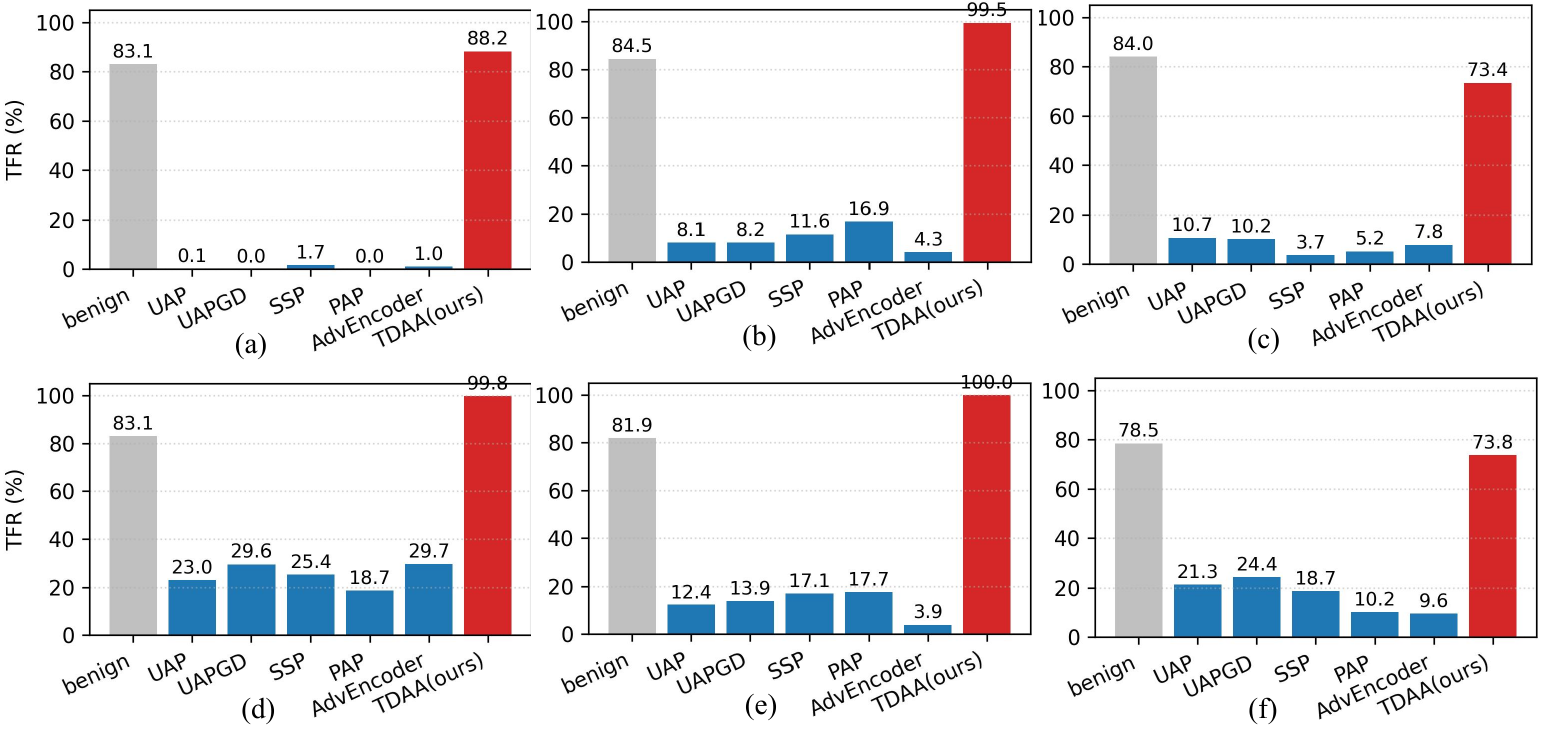} 
\caption{Comparison of TDAA against other attacks. `benign' indicates the baseline accuracy under no attack. Encoders a-f are pre-trained with BYOL, MoCo v2+, MoCo v3, NNCLR, SimCLR, and W-MSE, respectively.}
\label{compare}
\end{figure*}

\begin{table*}[]
\caption{The TFRs (\%) of transferability study. }
\label{transferability}
\centering
\begin{tabular}{cccccccccc|c}
\hline
Source & WMSE  & SimCLR & MoCo2+ & MoCo3 & NNCLR & ReSSL & SupCon & SwAV   & VibCreg & Average   \\ \hline
SimCLR & 84.60 & 100.00 & 98.18  & 99.00 & 63.12 & 80.28 & 99.77  & 99.87  & 19.59   & 82.71 \\
SupCon & 54.49 & 84.82  & 70.96  & 86.74 & 99.27 & 65.86 & 100.00 & 88.11  & 59.38   & 78.85 \\
MoCo2+ & 42.52 & 79.80  & 100.00 & 99.11 & 99.43 & 97.04 & 93.68  & 52.86  & 68.10   & 81.39 \\
SwAV   & 47.58 & 93.30  & 53.49  & 57.63 & 20.66 & 90.77 & 83.95  & 100.00 & 3.61    & 61.22 \\ \hline
\end{tabular}

\end{table*}
\subsection{Comparison Experiments}
To further evaluate the effectiveness of our method, we compare it with other DAE methods, including AdvEncoder~\cite{dae}, PAP~\cite{PAP}, UAP~\cite{uap}, UAPGD~\cite{uapgd}, and SSP~\cite{ssp}.






For implementation, we use CIFAR10 as the attacker dataset for all comparison methods. We select six SSL methods as the victim models, while
the STL10 dataset is used for the downstream task. The results, shown in Table~\ref{compare},
demonstrate that TDAA outperforms existing DAA methods across all experimental settings, even without knowledge of the downstream dataset. Notably, TDAA requires only a pre-trained encoder without a classification head to execute the attack, achieving an average TFR of 89\%. This demonstrates superior performance over existing approaches.

\subsection{Transferability Attack Experiments}
This section illustrates the transferability of our proposed method. We aim to demonstrate the effectiveness of our attack in the challenging scenario where our adversarial noise generator is trained on one specific SSL method and evaluated on different SSL methods. Specifically, we employ encoders pre-trained with SimCLR, SupCon, MoCo v2+ and SwAV as source models to attack a set of target encoders. All models are based on a ResNet-18 backbone, and experiments are performed on the CIFAR10 dataset. As demonstrated in Table~\ref{transferability}, the results indicate that TDAA successfully executes transfer attacks without requiring prior knowledge of the target's underlying SSL method. 
Instead of creating noise overfitted to a single model's specifics, our generator learns a more fundamental and robust transformation to disrupt the common semantic features learned by various SSL methods. This makes the generated perturbations inherently transferable.

\begin{table}[!t]
\caption{TFR (\%) of TDAA under different $\alpha$.}
\label{diff_alpha}
\centering
\begin{tabular}{cccc}
\hline
$\alpha$        & CIFAR10 & STL10  & ANIMALS10 \\ \hline
$\alpha=2$      & 100.00  & 100.00 & 98.25     \\
$\alpha=5$      & 100.00  & 100.00 & 100.00    \\
$\alpha=\infty$ & 100.00  & 100.00 & 100.00    \\ \hline
\end{tabular}
\end{table}

\subsection{Ablation Study}
\subsubsection{Impact of $\alpha$.}
\begin{figure}[!t]
    \centering
    \setlength{\tabcolsep}{4pt} 
    \renewcommand{\arraystretch}{1.5} 
    
    \begin{tabular}{c c}
        \raisebox{1.2\height}{\rotatebox[origin=c]{90}{Benign}} &
        \includegraphics[width=\columnwidth]{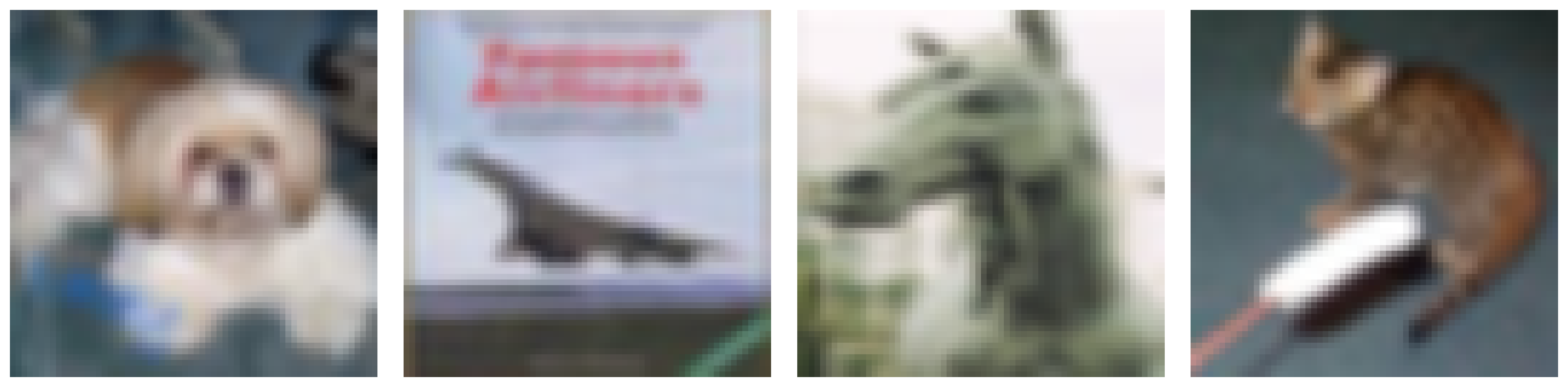} \\
        \raisebox{1.2\height}{\rotatebox[origin=c]{90}{\textbf{$\alpha=2$}}} &
        \includegraphics[width=\columnwidth]{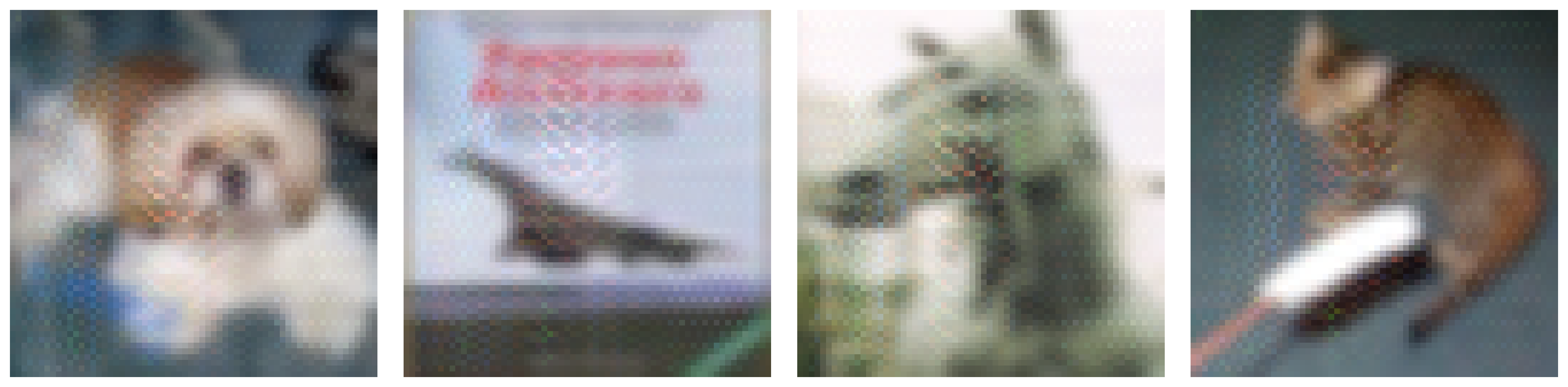} \\
        \raisebox{1.2\height}{\rotatebox[origin=c]{90}{\textbf{$\alpha=5$}}} &
        \includegraphics[width=\columnwidth]{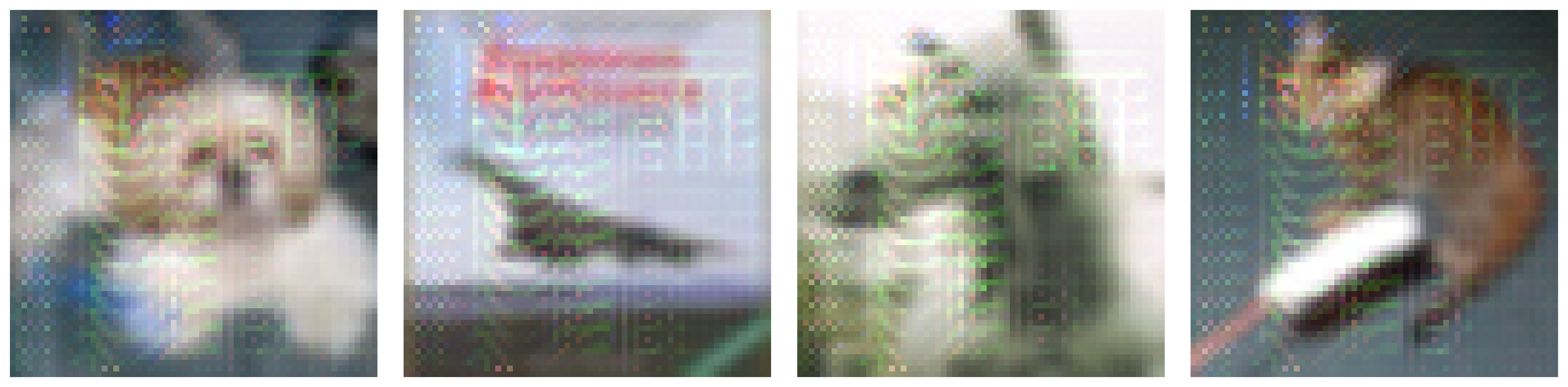} \\
        \raisebox{1.2\height}{\rotatebox[origin=c]{90}{\textbf{$\alpha=\infty$}}} &
        \includegraphics[width=\columnwidth]{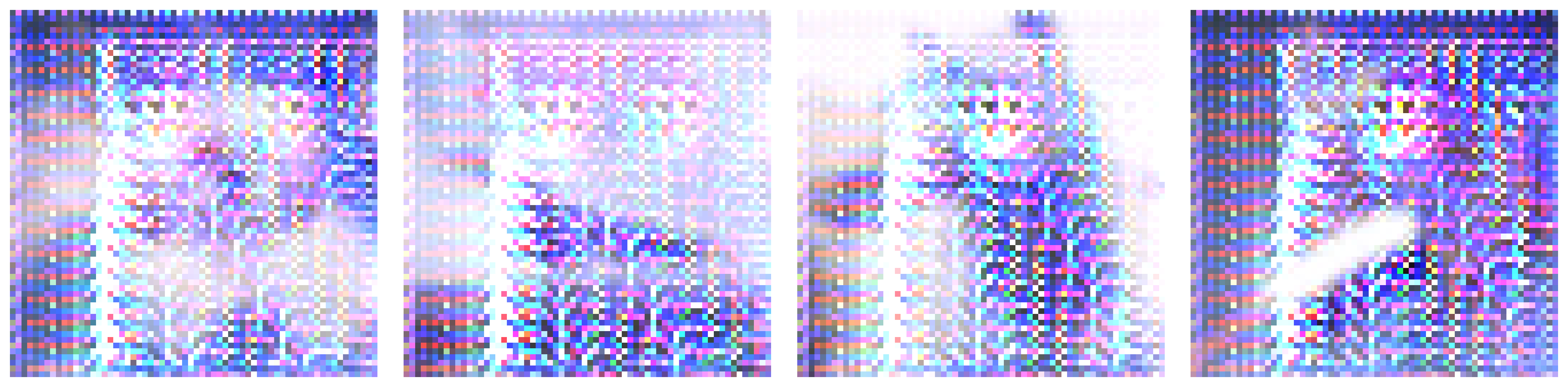}
    \end{tabular}
    \caption{Comparison of adversarial examples and original samples under different $\alpha$ values.}
    \vspace{-15pt}
    \label{ablation_quality}
\end{figure}

The composite loss function in Eq.~\eqref{eq:all_loss} integrates $L_{adv}$ for adversarial efficacy optimization and $L_{con}$ for visual imperceptibility. The weighting coefficient $\alpha$ governs the balance between these two objectives during gradient descent. In this subsection, we conducted experiments with: (1) $\alpha=2$, (2) $\alpha=5$, and (3) $L_{con}$ removal ($\alpha\to\infty$) to evaluate the visual and quantitative results of the adversarial examples. The experimental results are shown in Figure~\ref{ablation_quality} and Table~\ref{diff_alpha}.

It can be observed that as \(\alpha\) increases, the TFR improves, while the visual quality of the DAEs deteriorates. When \(\alpha = 2\), the generated DAEs achieve a good balance between attack success and perceptual similarity, which makes it our empirically recommended setting. Increasing \(\alpha\) to 5 results in noticeable perturbations. When $L_{con}$ is completely removed, the generated DAEs become visually different from the original samples. These results indicate the importance of $L_{con}$ in preserving the visual quality of the DAEs.

\subsubsection{Impact of Epoch Numbers.}
\begin{figure}[t]
\centering
\includegraphics[width=\columnwidth]{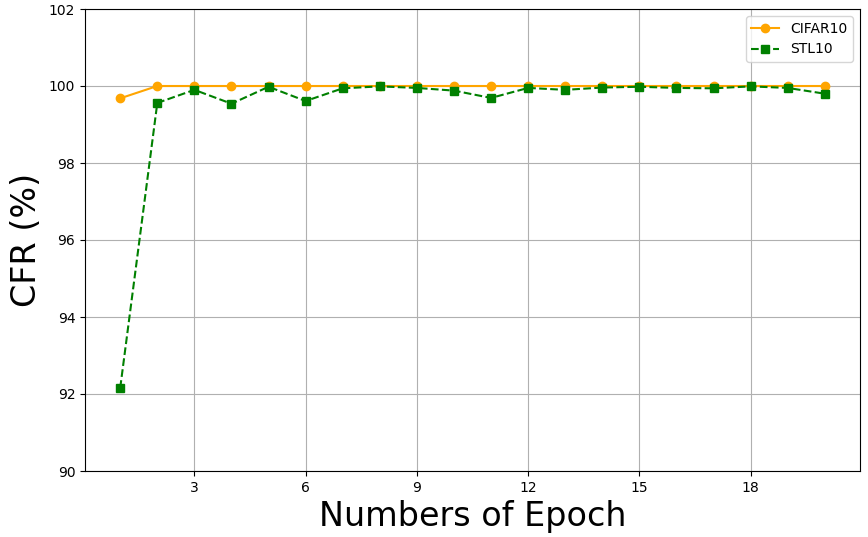} 
\caption{Performance of different epochs.}
\label{diff_epoch}
\end{figure}
We investigated the impact of different epochs on the generator's performance. Using NNCLR as the victim model and CIFAR10 as the attacker's dataset, we conducted experiments on downstream tasks using both CIFAR10 and STL10 across various epochs. The results are shown in Figure~\ref{diff_epoch}. These results clearly demonstrate that even at lower epochs, TDAA maintains strong attack effectiveness, which further highlights the vulnerability of pre-trained encoders.

\subsubsection{Impact of Threat Image.}

    
        
        
    

\begin{figure*}[t]
\centering

\includegraphics[width=\textwidth]{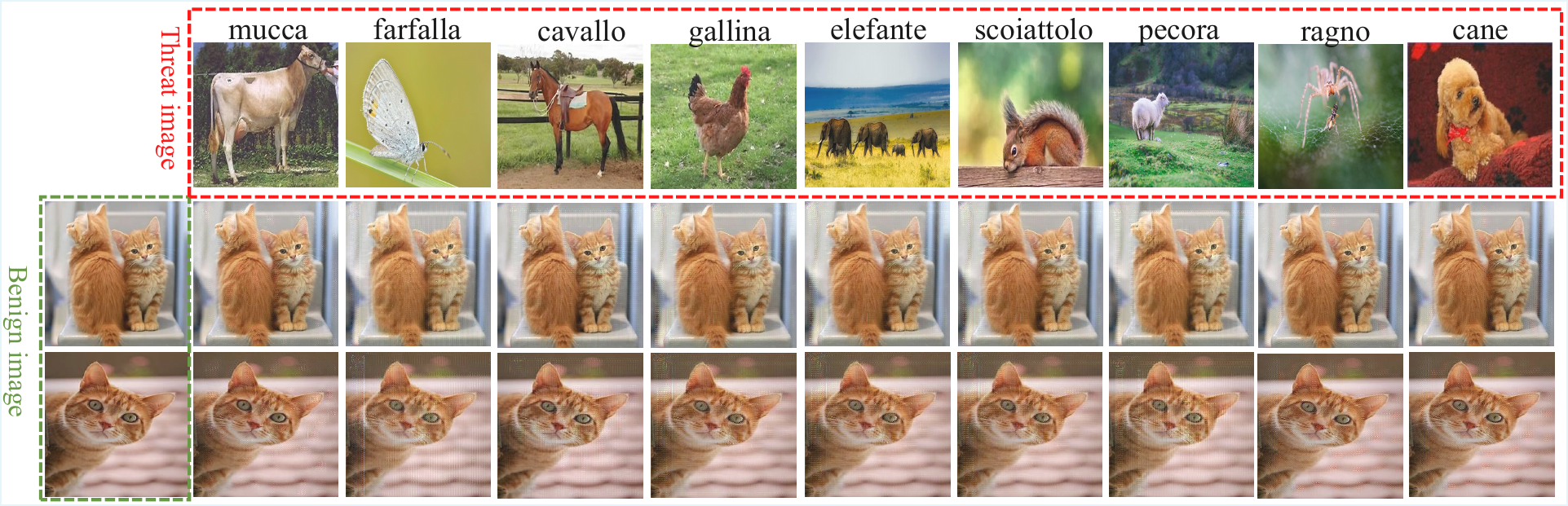} 
\caption{Different target images. The first row consists of different threat images, and the second and third rows consist of adversarial examples corresponding to each threat image.}
\label{muti_tar}
\end{figure*}

\begin{table}[]
\caption{The TFRs (\%) of TDAA under different threat images. }
\label{TFR_diff_tar}
\centering
\begin{tabular}{cccc|c}
\hline
\diagbox{$x_t$}{$D_d$}      & CIFAR10 & STL10  & ANIMALS10 & Average \\ \hline
cane       & 100.00  & 98.73  & 100.00    & 99.58   \\
cavallo    & 99.79   & 94.71  & 82.51     & 92.34   \\
elefante   & 87.97   & 88.22  & 92.11     & 89.43   \\
farfalla   & 85.00   & 85.60  & 80.46     & 83.69   \\
gallina    & 85.80   & 86.19  & 80.62     & 84.20   \\
mucca      & 94.88   & 85.72  & 87.35     & 89.32   \\
pecora     & 100.00  & 100.00 & 100.00    & 100.00  \\
ragno      & 98.02   & 73.47  & 97.92     & 89.80   \\
scoiattolo & 99.39   & 89.94  & 89.17     & 92.83   \\ \hline
\end{tabular}
\end{table}
In previous experiments, we used a single threat image. To eliminate randomness in threat image selection, we introduced nine additional threat images for the experiments. These images were randomly selected from the ANIMALS10 dataset. In the experiments, the ANIMALS10 dataset served as the attacker's dataset, while CIFAR10 and STL10 were selected as the downstream datasets. The SupCon model acted as the victim model in the experiments. The experimental results are presented in Table~\ref{TFR_diff_tar}, and the selected threat images are shown in Figure~\ref{muti_tar}.

From Table~\ref{TFR_diff_tar}, it is evident that regardless of the chosen threat image, a high TFR is consistently achieved, which supports effective attack performance while maintaining satisfactory visual quality. This indicates that TDAA has minimal requirements for threat image selection, and almost any image can serve as a target.

\subsubsection{Impact of Encoder's Backbone}
To verify that the efficacy of TDAA is not constrained to a specific architecture, we extended our evaluation from the previously used ResNet-18 to an advanced backbone, Vision Transformer (ViT)~\cite{vit}. Specifically, in this experiment, the ViT encoder was pre-trained on CIFAR10 via the DINO framework. We then assessed its vulnerability to TDAA on the CIFAR10, STL10, and ANIMALS10 datasets. The results, summarized in Table~\ref{vit}, demonstrate that TDAA maintains an exceptionally high TFR against the ViT model, confirming its attack effectiveness. Experiments illustrate that TDAA is agnostic to the underlying encoder architecture.

\begin{table}[!t]
\caption{The TFRs (\%) of different distance criteria.}
\label{diff_cri}
\resizebox{0.48\textwidth}{!}{
\begin{tabular}{cccccccc}
\hline
Dataset                  & criterion & NNCLR  & MoCo2+ & MoCo3  & ReSSL  & SupCon & SimCLR \\ \hline
\multirow{3}{*}{CIFAR10} & L2        & 100.00 & 100.00 & 99.98  & 100.00 & 100.00 & 100.00 \\
                         & COS       & 100.00 & 0.00   & 100.00 & 0.00   & 10.92  & 100.00 \\
                         & InfoNCE   & 11.02  & 11.29  & 10.89  & 11.19  & 10.02  & 11.34  \\
\multirow{3}{*}{STL10}   & L2        & 99.80  & 99.48  & 73.39  & 98.93  & 100.00 & 100.00 \\
                         & COS       & 0.00   & 0.00   & 100.00 & 80.09  & 70.92  & 100.00 \\
                         & InfoNCE   & 10.30  & 12.02  & 9.92   & 10.33  & 9.55   & 9.52   \\ \hline
\end{tabular}

}

\end{table}

\begin{table}[!t]
\caption{The TFRs (\%) against the ViT-based encoder.}
\label{vit}
\resizebox{0.48\textwidth}{!}{
\begin{tabular}{cccc|c}
\hline
\diagbox{$D_a$}{$D_d$} & CIFAR10 & STL10 & ANIMALS10 & Average   \\ \hline
CIFAR10   & 99.80    & 99.86  & 52.53     & 84.06 \\
STL10     & 91.71    & 94.92  & 56.03     & 80.89 \\
ANIMALS10  & 91.04    & 99.05  & 86.73     & 92.27 \\ \hline
\end{tabular}
}

\end{table}
\subsubsection{Impact of the Feature Distance Criterion.}
In TDAA, we utilize the \(L_2\) distance to measure the distance between features. To assess the performance of other metrics, we conducted experiments using cosine distance and InfoNCE~\cite{nce} distance as alternative measures. The attacker's dataset is CIFAR10, and the downstream tasks use CIFAR10 and STL10. The results are shown in Table~\ref{diff_cri}. 

It is evident that using the \(L_2\) distance yields the best results. Cosine distance exhibits significant instability, with the TFR dropping to very low levels, even reaching 0\% in some settings. InfoNCE consistently shows a low attack success rate. 
Since cosine similarity only considers the angle between vectors and not their magnitude, it may fail to capture subtle adversarial perturbations that affect feature magnitudes.
Although InfoNCE has performed well in previous work~\cite{dae}, it underperforms in TDAA. This is primarily due to a mismatch between its training objective and the requirements of TDAA. InfoNCE aims to maximize the similarity between positive pairs while minimizing the similarity between the anchor and a set of negative samples. However, TDAA targets a fundamentally different goal: identifying all samples as the same target, implying the lack of explicit negative samples. In previous adversarial tasks, it was sufficient to induce output inconsistency between benign and adversarial examples, where pushing apart negative samples helped improve attack effectiveness. As a result, InfoNCE was well-suited for such tasks. In contrast, its reliance on negative sampling renders it less effective in our setting.

\begin{table}[!t]
\caption{Comparison of the TFRs (\%) between fixed noise and noise generator.}
\label{fix_noise}
\centering
\resizebox{\columnwidth}{!}{
\begin{tabular}{ccccccc}
\hline
\multicolumn{1}{l}{} & \multicolumn{2}{c}{CIFAR10} & \multicolumn{2}{c}{STL10} & \multicolumn{2}{c}{ANIMALS10} \\
Victim               & fixed noise     & TDAA      & fixed noise    & TDAA     & fixed noise      & TDAA       \\ \hline
BYOL                 & 1.30            & 100.00    & 0.18           & 99.48    & 0.00             & 99.86      \\
NNCLR                & 45.23           & 100.00    & 95.20          & 100.00   & 0.00             & 100.00     \\
ReSSL                & 2.08            & 100.00    & 80.24          & 100.00   & 0.22             & 100.00     \\
SimCLR               & 89.29           & 99.83     & 93.79          & 99.96    & 22.92            & 100.00     \\
SupCon               & 88.07           & 100.00    & 71.75          & 100.00   & 1.71             & 99.95      \\
SwAV                 & 15.25           & 100.00    & 91.52          & 100.00   & 36.45            & 100.00     \\
VibCreg              & 3.34            & 100.00    & 99.88          & 100.00   & 2.64             & 100.00     \\ \hline
Average                  & 34.94           & 99.98     & 76.08          & 99.92    & 9.13             & 99.97      \\ \hline
\end{tabular}
}
\end{table}

\subsubsection{Impact of Personalized Generation Paradigm}
In addition to the difference in attack targets compared to previous DAE, TDAA introduces a new DAE generation paradigm that generates example-specific adversarial perturbations. In contrast, previous DAA methods assume that all samples share a single perturbation and optimize this noise directly during training, rather than optimizing a noise generation network. To investigate whether this change improves attack performance, we designed an experiment where the noise generator was replaced with direct noise optimization. All other experimental settings were kept unchanged. For this experiment, we used ANIMALS10 as the attacker's dataset, and CIFAR10, STL10, and ANIMALS10 as downstream datasets. The results are summarized in Table~\ref{fix_noise}.\par

The findings reveal that directly optimizing noise yields significantly worse results. In some cases, the TFR values fall below 20\%, much lower than those achieved with the noise generator. This discrepancy arises because achieving targeted attack objectives requires customized noise for each benign sample. While the noise generator can produce personalized noise for each sample, using shared noise across all samples fails to meet this requirement effectively.

\begin{table}[!t]
\caption{The TFRs (\%) of TDAA on fine-tuned encoders. }
\centering
\label{TFR_hot_encoder}
\begin{tabular}{cccc|c}
\hline
Victim  & CIFAR10 & STL10  & ANIMALS10 & Average \\ \hline
SupCon  & 99.96   & 100.00 & 100.00    & 99.99   \\
SimCLR  & 99.95   & 99.93  & 99.43     & 99.77   \\
NNCLR   & 97.71   & 99.98  & 100.00    & 99.23   \\
ReSSL   & 70.65   & 84.99  & 97.14     & 84.26   \\
SwAV    & 99.93   & 100.00 & 99.62     & 99.85   \\
VibCreg & 100.00  & 100.00 & 100.00    & 100.00  \\ \hline
\end{tabular}
\end{table}

\begin{table}[!t]
\caption{Performance of TDAA with CLIP.}
\label{clip}
\resizebox{\columnwidth}{!}{
\begin{tabular}{ccccc|c}
\hline
$D_{d} $                  & Metric          & CIFAR10 & STL10  & ANIMALS10 & Average \\ \hline
\multirow{2}{*}{CIFAR10}   & ATA$\downarrow$ & 10.21   & 10.31  & 32.54     & 17.69   \\
                           & TFR$\uparrow$   & 99.98   & 99.58  & 67.67     & 89.08   \\ \hline
\multirow{2}{*}{STL10}     & ATA$\downarrow$ & 10.21   & 9.99   & 9.60      & 9.93    \\
                           & TFR$\uparrow$   & 100.00  & 100.00 & 97.18     & 99.06   \\ \hline
\multirow{2}{*}{ANIMALS10} & ATA$\downarrow$ & 10.21   & 10.05  & 7.35      & 9.20    \\
                           & TFR$\uparrow$   & 99.97   & 99.80  & 99.41     & 99.73   \\ \hline
\end{tabular}
}
\end{table}

\subsection{Fine-Tuning Encoders}
In previous experiments, during the downstream training phase, the encoder's parameters were frozen and remained unchanged, while only the downstream heads were optimized. However, in real-world scenarios, downstream users often fine-tune the encoder's parameters, which leads to discrepancies between the attacker's encoder and the final encoder used by the user. To examine whether such differences affect the effectiveness of TDAA, we designed the following experiment.

In this experimental setting, the encoder's parameters were not frozen during downstream model training. Instead, the encoder was fine-tuned alongside the training of the downstream heads. After training, TDAA was applied to attack the fine-tuned model. For the downstream task, we selected STL10 as the dataset, while the attacker's datasets included CIFAR10, STL10, and ANIMALS10. The experimental results are presented in Table~\ref{TFR_hot_encoder}.

From the results, it can be observed that even when users fine-tune the encoder's parameters, TDAA still demonstrates promising attack performance. This is because the encoder, pre-trained on a large-scale dataset, has parameters that are already close to convergence. During training, the updates primarily focus on the downstream model, and the encoder's parameters undergo minimal changes after the model converges. As a result, TDAA remains highly effective despite the fine-tuning.

\subsection{Experiments on Multimodal Pre-Trained Encoders}
The above experiments validate the effectiveness of our method on vision-only pre-trained models. To further investigate the performance of TDAA on multimodal models and larger-scale architectures, we conducted experiments on CLIP~\cite{clip}. Specifically, we evaluated the performance of TDAA on the CLIP model with a ViT-B/32 backbone using CIFAR10, STL10, and ANIMALS10, and the experimental results are reported in Table~\ref{clip}. From the results, we observe that TDAA consistently achieves a high TFR under this attack scenario, which demonstrates strong attack effectiveness even against multimodal models. This robustness demonstrates that our method is a general attack paradigm, which enables the exploration of latent vulnerabilities in pre-trained models, instead of tailoring attacks to a specific backbone.

\subsection{Extension to Other Downstream Tasks}
To further evaluate the performance of TDAA on other downstream tasks, we conducted experiments on the retrieval task. Specifically, the encoder was pre-trained on CIFAR10, while both the attacker’s surrogate dataset and the downstream task datasets are selected from CIFAR10, STL10, and ANIMALS10. In this experiment, the training set serves as the retrieval gallery, and the test set is used as the query set. Table~\ref{retrieval} reports the Top-10 TFR of different scenarios.

The results show that TDAA maintains a high TFR on the retrieval task, demonstrating strong attack effectiveness across downstream tasks. This is because TDAA is not designed for a specific downstream task, but rather targets the encoder itself, enabling it to achieve consistently high attack effectiveness across different downstream tasks.
\begin{table*}[]
\caption{The TFRs (\%) of TDAA on the retrieval task.\( D_{a} \) represents the dataset used by the attacker, while \( D_{d} \) denotes the dataset used for downstream tasks.}
\label{retrieval}
\resizebox{\textwidth}{!}{
\begin{tabular}{ccllllllllll|c}
\hline
$D_a$                   & $D_d$  & \multicolumn{1}{c}{BYOL}   & \multicolumn{1}{c}{W-MSE} & \multicolumn{1}{c}{SimCLR} & \multicolumn{1}{c}{MoCo2+} & \multicolumn{1}{c}{MoCo3} & \multicolumn{1}{c}{NNCLR} & \multicolumn{1}{c}{RESSL} & \multicolumn{1}{c}{SupCon} & \multicolumn{1}{c}{SwAV} & \multicolumn{1}{c|}{VibCreg} & AVG   \\ \hline
\multirow{3}{*}{CIFAR10}   & CIFAR10   & \multicolumn{1}{c}{100.00} & 99.36                     & 100.00                     & 100.00                     & \multicolumn{1}{c}{99.95} & 100.00                    & 100.00                    & 100.00                     & 100.00                   & 100.00                       & 99.93 \\
                           & STL10     & 88.87                      & 88.98                     & 100.00                     & 99.95                      & 92.41                     & 99.55                     & 100.00                    & 100.00                     & 100.00                   & 99.99                        & 96.98 \\
                           & ANIMALS10 & 48.79                      & 33.54                     & 99.98                      & 99.92                      & 75.92                     & 50.29                     & 99.98                     & 100.00                     & 100.00                   & 88.02                        & 79.64 \\ \hline
\multirow{3}{*}{STL10}     & CIFAR10   & 100.00                     & 84.50                     & 100.00                     & 99.98                      & \multicolumn{1}{c}{98.95} & 100.00                    & 100.00                    & 100.00                     & 100.00                   & 100.00                       & 98.34 \\
                           & STL10     & 99.99                      & 96.92                     & 100.00                     & 100.00                     & 94.95                     & 100.00                    & 100.00                    & 100.00                     & 100.00                   & 100.00                       & 99.19 \\
                           & ANIMALS10 & 98.89                      & 68.91                     & 100.00                     & 83.74                      & 92.61                     & 99.89                     & 100.00                    & 100.00                     & 100.00                   & 99.97                        & 94.40 \\ \hline
\multirow{3}{*}{ANIMALS10} & CIFAR10   & 99.99                      & 77.83                     & 100.00                     & 100.00                     & \multicolumn{1}{c}{97.55} & 100.00                    & 100.00                    & 100.00                     & 100.00                   & 100.00                       & 97.54 \\
                           & STL10     & 99.95                      & 87.37                     & 100.00                     & 100.00                     & 81.01                     & 100.00                    & 100.00                    & 100.00                     & 100.00                   & 100.00                       & 96.83 \\
                           & ANIMALS10 & 99.95                      & 96.53                     & 100.00                     & 100.00                     & 96.60                     & 100.00                    & 100.00                    & 100.00                     & 100.00                   & 100.00                       & 99.31 \\ \hline
\end{tabular}
}
\end{table*}
\section{Conclusions}

In this paper, we propose a novel and stricter threat model to assess the vulnerability of pre-trained encoder-based methods. In contrast to previous DAA approaches, we focus on achieving Targeted DAA. Specifically, we introduce a novel example-specific DAE generation paradigm that tailors adversarial perturbations to individual samples, which ensures that the downstream model generates the expected prediction. Extensive experimental results highlight the vulnerabilities of pre-trained encoder-based methods, even under the proposed stricter threat model. Our findings underscore the need for improved security measures and rigorous evaluations for pre-trained models in future work.
\bibliographystyle{ieeetr}
\bibliography{TDSC/ref}

\vfill

\end{document}